\documentclass[runningheads]{llncs}
\usepackage{xcolor}
\usepackage{caption}
\usepackage{subcaption}
\usepackage{changes} 
\usepackage{booktabs}
\usepackage{CJKutf8} 
\usepackage{tcolorbox}
\usepackage{multirow}
\usepackage[numbers]{natbib}  

\makeatletter
\renewcommand\@biblabel[1]{#1.}
\makeatother

\usepackage{amsmath}
\usepackage[T1]{fontenc}

\usepackage[colorlinks=true, citecolor=blue]{hyperref}

\begin{document}
\title{Chinese Metaphor Recognition Using a Multi-stage Prompting Large Language Model}
%
%
\author{Jie Wang \and Jin Wang \and Xuejie Zhang}
\institute{School of Information Science and Engineering\\
Yunnan University\\
\email{wangjie\_qpqj@stu.ynu.edu.cn}, \email{\{wangjin, xjzhang\}@ynu.edu.cn}
}
\maketitle              
\begin{abstract}
Metaphors are common in everyday language, and the identification and understanding of metaphors are facilitated by models to achieve a better understanding of the text. Metaphors are mainly identified and generated by pre-trained models in existing research, but situations, where tenors or vehicles are not included in the metaphor, cannot be handled. The problem can be effectively solved by using Large Language Models (LLMs), but significant room for exploration remains in this early-stage research area. A multi-stage generative heuristic-enhanced prompt framework is proposed in this study to enhance the ability of LLMs to recognize tenors, vehicles, and grounds in Chinese metaphors. In the first stage, a small model is trained to obtain the required confidence score for answer candidate generation. In the second stage, questions are clustered and sampled according to specific rules. Finally, the heuristic-enhanced prompt needed is formed by combining the generated answer candidates and demonstrations. The proposed model achieved 3rd place in Track 1 of Subtask 1, 1st place in Track 2 of Subtask 1, and 1st place in both tracks of Subtask 2 at the NLPCC-2024 Shared Task 9.

\keywords{Chinese metaphor generation \and Multi-stage prompting \and  Large language model \and DeBERTa.}
\end{abstract}

\section{Introduction}
Metaphor is an essential tool for reasoning and linguistic expression, the task is a crucial step toward the generation of human-like language. When machines learn the human habit of creating metaphors, the first step is to identify the tenors and vehicles in human-created metaphors. The tenor represents the subject, while the vehicle represents the comparative element. 

Currently, proposed methods mainly focus on using neural networks, such as BiLSTM \cite{bizzoni-ghanimifard-2018-bigrams,mao-etal-2019-end,pramanick-etal-2018-lstm} and BERT \cite{su-etal-2020-deepmet, chen-etal-2020-go,zhang-liu-2022-metaphor} to identify tenors and vehicles in metaphors. Although the models used are different, they essentially work by identifying the probability that each token in the metaphor belongs to these two components to obtain tenors (\begin{CJK*}{UTF8}{gbsn}本体\end{CJK*}) and vehicles (\begin{CJK*}{UTF8}{gbsn}喻体\end{CJK*}). However, this approach is only practical for metaphors that are similes. The simile is a particular type of metaphor that compares tenors and vehicles of different categories using comparator words such as "like", "as", or "than" and tenors and vehicles will appear directly in the simile. For example, in the phrase "\begin{CJK*}{UTF8}{gbsn}闪电像火蛇\end{CJK*}" (lightning like a fire snake), "\begin{CJK*}{UTF8}{gbsn}闪电\end{CJK*}" (lightning) is the tenor, and "\begin{CJK*}{UTF8}{gbsn}火蛇\end{CJK*}" (fire snake) is the vehicle.

However, in many metaphors, tenors, and vehicles are not obvious. For example, in the phrase "\begin{CJK*}{UTF8}{gbsn}飞流直下三千尺，疑是银河落九天\end{CJK*}" (flying straight down three thousand feet, suspected to be the Milky Way falling into the sky), the tenor is "\begin{CJK*}{UTF8}{gbsn}瀑布\end{CJK*}" (waterfall), but "\begin{CJK*}{UTF8}{gbsn}瀑布\end{CJK*}" (waterfall) does not directly appear in the metaphor. Therefore, correctly identifying tenors and vehicles in metaphors is a significant step for metaphor generation tasks. In response to this situation, some studies have used pre-trained models for fine-tuning to generate tenors and vehicles in metaphors, such as T5 \cite{stowe-etal-2021-exploring, li2024finding} and BART \cite{chakrabarty2020generating,chakrabarty-etal-2021-mermaid, stowe2021metaphor}. However, this requires a considerable amount of computational resources and data, and these methods make it difficult to explain the reasoning behind metaphorical/literal judgments. Some studies \cite{tian-etal-2024-theory, comsa2022miqa, tong2024metaphor, Shao2024CMDAGAC} have started using LLMs to identify metaphors and generate metaphors, constructing a series of examples through the use of metaphor theory, but manually constructed examples are expensive and may not necessarily be understandable to LLMs.

Considering the above situation, a multi-stage method is proposed to prompt large language models to correctly identify tenors, vehicles, and grounds (\begin{CJK*}{UTF8}{gbsn}共性/喻意\end{CJK*}) in metaphors. In the first stage, answer candidates are generated by the DeBERTa model, which can help LLMs choose from multiple possible answers and prioritize those with higher confidence. In the second stage, demonstrations generated by LLMs do not require manual generation and can teach the underlying reasoning logic of metaphor recognition to the model without using metaphor theory.

The proposed system participated in the NLPCC-2024 Shared Task 9 \footnote{https://github.com/xingweiqu/NLPCC-2024-Shared-Task-9}. This shared task consists of two subtasks, each including two evaluation tracks. Our method achieved an accuracy of 0.959 in subtask1\_track1, 0.979 in subtask1\_track2, 0.951 in subtask2\_track1, and 0.941 in subtask2\_track2. Except for ranking third in subtask1\_track1, it ranked first in all other subtasks.

The rest of this paper is organized as follows. The related work on metaphor generation is introduced in Section 2. A detailed description of the proposed system and model is provided in Section 3. The experiment and results are discussed in Section 4. Finally, Section 5 presents the conclusion.

\section{Related Work}
\subsubsection{Automated Metaphor Identification.} Most current work \cite{liu2018neural,gao-etal-2018-neural,bizzoni-ghanimifard-2018-bigrams,mao-etal-2019-end,pramanick-etal-2018-lstm} treats metaphor identification as a sequence labeling task using the BiLSTM architecture, outputting metaphorical label sequences for input word sequences (typically sentences). With the introduction of the Transformer, various models based on the Transformer, such as Bidirectional Encoder Representations from Transformers (BERT) \cite{devlin2018bert}, RoBERTa \cite{liu2019roberta}, DeBERTa \cite{he2020deberta}, etc., have been widely used in various tasks of natural language processing and have achieved amazing results. Many recent studies have used pre-trained contextual language models, such as BERT \cite{chen-etal-2020-go,dankers-etal-2020-neighbourly} and the variant model RoBERTa \cite{gong-etal-2020-illinimet, li2024finding} of BERT, achieving significant results. In this paper, the DeBERTa model is used to construct answer candidates in the first stage, which is used to provide recommendations to LLMs.
\subsubsection{Metaphor Recognition with LLMs.} 
Neidlein et al. \cite{neidlein-etal-2020-analysis} proposed that the success of most metaphor recognition systems currently using pre-trained models is due to optimizing the disambiguation of conventionalized, metaphoric word senses for specific words instead of modeling general properties of metaphors. Recently, some studies \cite{tian-etal-2024-theory, comsa2022miqa, tong2024metaphor, Shao2024CMDAGAC, lee2022chinese} have started using LLMs to identify metaphors and generate metaphors, constructing a series of examples through the use of metaphor theory, directing the LLM to incrementally generate the reasoning process for metaphor understanding through dialogue interactions.  

\subsubsection{Instruction Tuning.}
By using Chain-of-Thought (CoT) prompting techniques, LLMs can be guided to decompose multi-step problems into intermediate steps before generating answers, which can improve their performance in complex reasoning tasks. CoT prompting can be classified into two major paradigms: Zero-Shot-CoT \cite{kojima2022large} and Manual-CoT \cite{wei2022chain}. Manual-CoT often achieves better results than Zero-Shot-CoT, but the cost of handmade instructions or prompts is expensive. In this paper, we use the Auto-CoT \cite{zhang2022automatic} paradigm to construct demos with questions and reasoning in chains automatically.
\section{Generative Heuristic-enhanced Prompt Framework Method}
\begin{figure}[t!]
\centering 
\includegraphics[height=6cm]{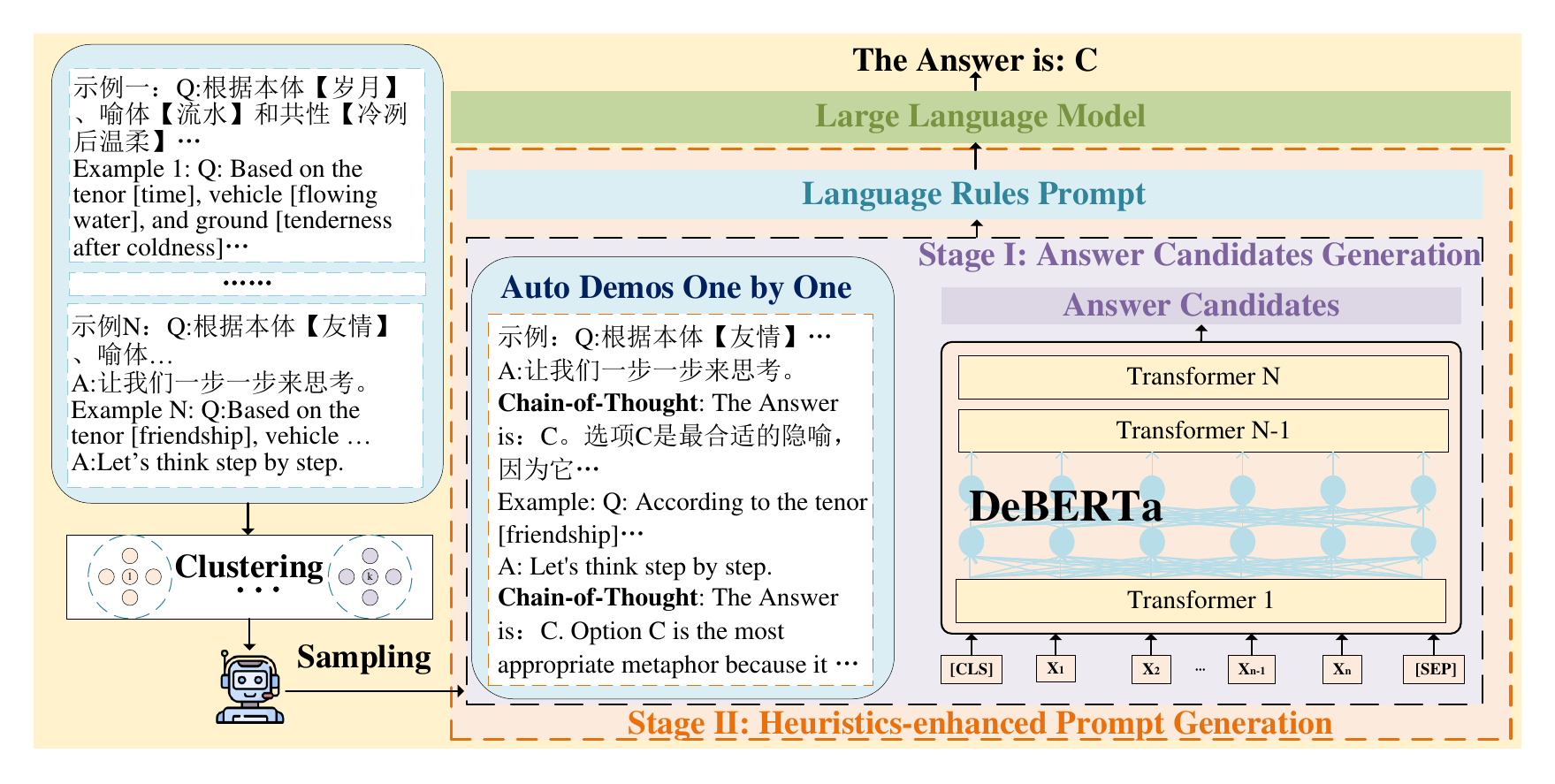}
\caption{The proposed approach consists of two stages: answer candidates generation and heuristic-enhanced prompt generation. In the first stage, the DeBERTa model generates a candidate list in brackets based on options and confidence scores. Then, the demonstrations are combined with the candidate list to form a heuristic-enhanced prompt.} 
\label{sample-figure}
\vspace{-2.0em}
\end{figure}

\subsection{Task Description}
NLPCC-2024 Shared Task 9 uses machine learning techniques to generate Chinese metaphors by effectively identifying the ground or vehicle in the metaphoric relation. It is divided into two subtasks \cite{Shao2024CMDAGAC,qu2024overviewnlpcc2024shared}:
\begin{itemize}
    \item[$\bullet$] Subtask 1. Metaphor Generation involves creating a metaphor from a provided tuple consisting of TENOR, GROUND, and VEHICLE. The goal here is to synthesize a metaphor that aptly connects the subject (TENOR) with the object (VEHICLE), guided by the concept of the GROUND.
    \item[$\bullet$] Subtask 2. Metaphor Components Identification, aimed at extracting the TENORS, GROUNDS, and VEHICLES from a symbolic sentence. This component requires identifying metaphor elements that correspond to the specified grounds.
\end{itemize}

\subsection{Stage I: Answer Candidates Generation}
Each sequence input to DeBERTa includes one question ${Q}$ and a list of options ${A}$. Each question ${Q}$ is represented as ${Q}=[q_1,q_2,\dots,q_n]$, each option list is represented as ${A}=[A_1,A_2,A_3,A_4]$, where each option $A_i$ is represented as $A_i=[a_{i1},a_{i2},\dots,a_{im}]$. Among them, $n$ represents the length of the problem text and $m$ represents the length of the option text. Combine question ${Q}$ with each option $A_i$ and insert a "[SEP]" mark between them. Embed words into the concatenated sequence and add a special tag "[CLS]" to the beginning of the sequence, to build an input sequence for classification tasks. For each question-option pairs $(Q,A_i)$, the embedded representation $T_i$, is calculated as follows:
\begin{equation}
T^\text{word}_i = \operatorname{Emb}([[\text{CLS}],q_1,q_2,...,q_n,[\text{SEP}],a_{i1},a_{i2},...,a_{im}])
\end{equation}
\begin{equation}
    T_i=T_i^\text{word}+T^\text{pos}+T^\text{seg}
\end{equation}
where $\operatorname{Emb}()$ is the embedding layer. $T_i^\text{word}$, $T^\mathrm{pos}$, and $T^{\mathrm{seg}}\in {R}^{(n+m+2)\times d}$ represent the word embedding, position embedding, and segment embedding, respectively, where $d$ is the dimension size of the embedding. During training, the cross-entropy as the loss function for optimizing the model:
\begin{equation}\mathcal{L}=-\frac1N\sum_{j=1}^N\sum_{i=1}^Cy_{ji}\log(p_{ji})\end{equation}
where $N$ is the number of samples, $C$ is the number of options, $y_{ji}$is the ground-truth label and $p_{ji}$ is the confidence score. For each option $A_i$, the corresponding confidence score $p_i$ is calculated using the DeBERTa model:
\begin{equation}
p_i=\text{softmax}(W\mathcal{M}^{DeBERTa}(T_i)+b)
\end{equation}
where $\mathcal{M}^{DeBERTa}$ represents the DeBERTa model, $W$ is the weight matrix, and $b$ is the bias term. Arrange the confidence scores of all options into a list $\mathbf{p}=[p_1,p_2,p_3,p_4]$. 
\subsection{Stage II: Heuristics-enhanced Prompt Generation}
\subsubsection{Question Clustering.} Firstly, a vector representation is computed for each question by DeBERTa.  Then, the question representations are processed by the k-means clustering algorithm to produce $k$ clusters of questions. The choice of $k$ here is defined using the elbow rule to determine the optimal number of clusters. Then, by using the phrase "Let's think step by step." to induce LLMs to generate reasoning chains for all problems \cite{su2017automatic}, the generated chain-of-thought reasoning is combined with the problems to obtain the dataset $\mathcal{D}$.
\subsubsection{Demonstration Sampling.}
To make the generated examples more accurate and easy to understand, examples where we finally choose the shortest encoding length for the question and the answer obtained when generating the chain-of-thought is consistent with the correct answer when sampling examples in dataset $\mathcal{D}$. After demonstration sampling for all the $k$ clusters, there will be $k$ constructed demonstrations $[e^{(1)},e^{(2)},\dots,e^{(k)}]$. The constructed demonstrations augment a test question $q^{test}$ for in-context learning. 
\subsubsection{Heuristic-enhanced Prompt Generation.}
After obtaining demonstrations $[e^{(1)},e^{(2)},\dots,e^{(k)}]$ and confidence score list $\mathbf{p}$, the heuristic prompt passed to the large model at the end is:
\begin{tcolorbox}[colback=white,colframe=black!75!black,title=]
\texttt{Demonstration: $[e^{(1)},e^{(2)},\dots,e^{(k)}]$ \textbackslash n \\
Q: $\textcolor{blue}{q^\text{test}}$ \textbackslash n \\
Answer candidates: $\textcolor{blue}{\mathbf{p}^\text{test}}$ \textbackslash n \\
A: "Let's think step by step."} 
\end{tcolorbox}
\noindent where the variables marked in blue will be substituted by specific testing inputs.

\section{Experiments}
\subsection{Dataset and Evaluation Metrics}
\subsubsection{Datasets.} Task 9 includes a training set of 34,463 metaphorical sentences with tenors, vehicles, and annotated grounds, and two validation sets of 500 sentences each, consistent with the test set format.
\subsubsection{Metrics.}
We use accuracy to evaluate the effectiveness of the model. For each task, two assessment tracks are provided:
\begin{itemize}
    \item[$\bullet$] Track 1: LLMs track. The Track encourages using large models to generate options directly. You can use your prompts, but please use a common prompt during the answer phase: "The answer is \{\}."
    \item[$\bullet$] Track 2: Rule-based track. The Track encourages using traditional language rules or machine learning-based methods to directly compare and draw conclusions about options A, B, C, and D.
\end{itemize}
\subsection{Implementation Details}
Qwen2-plus was chosen as the LLM in the response generation stage due to its ability to generate high-quality Chinese responses. First, a vector representation for each question is computed using DeBERTa. Then, the question representations are processed by the k-means clustering algorithm to produce three clusters of questions. The checkpoint used for DeBERTa is IDEA-CCNL/Erlangshen-DeBERTa-v2-710M-Chinese \footnote{https://huggingface.co/IDEA-CCNL/Erlangshen-DeBERTa-v2-710M-Chinese}. An AdamW optimizer with a learning rate of 2e-5 is used. Due to the limitation of computing resources, the batch size is set to 1. The weight decay is set to 0.01. The neural ranker is trained for 5 epochs.
\subsection{Empirical Results} 
\begin{table}[t!]
\vspace{-2.0em}
\centering
\caption{The results of subtask1 on the validation set}
\label{subtask1}
\begin{tabular}{ll}
\hline \toprule[1.5pt]
Methods                                                 & Accuracy       \\ \hline
Human's language rules                                  & 0.838          \\
DeBERTa-un\_finetuned                                   & 0.234          \\
DeBERTa-finetuned                                       & \textbf{0.97} \\ \hline
LLMs                                                     & 0.73           \\
prompt 1: DeBERTa-finetuned results                                       & 0.65           \\
prompt 2: Language Rules results                                 & 0.624          \\
prompt 3: Language Rules results \& DeBERTa results             & 0.592          \\
prompt 4: DeBERTa results \& LLMs results and reasons            & 0.576          \\
prompt 5: Language Rules results \& LLMs results and reasons & 0.564          \\
prompt 6: DeBERTa-un\_finetuned generate candidates              & 0.766          \\
prompt 7: DeBERTa-finetuned generate candidates                  & 0.964          \\ \hline
Our Method                                              & \textbf{0.98} \\ 
w/o candidate                                           & 0.906               \\
w/o demonstration                                       & 0.974         \\ \hline \toprule[1.5pt]
\end{tabular}
\vspace{-2.0em}
\end{table}

\begin{table}[t!]
\centering
\caption{Official Results}
\label{official_rusults}
\begin{tabular}{lllll}
\toprule[1.5pt]
Task                                  & Team Name         &Test  A    & Test B     & Average Score \\ \toprule[1.5pt]
\multirow{3}{*}{subtask1\_track1} & kangreen          & 98.8 & 98.0  & 98.4  \\
                                  & ShaunTheSheep     & 96.2 & 96    & 96.1  \\
                                  & \textbf{YNU-HPCC} & \textbf{96.6} & \textbf{95.2}  & \textbf{95.9}  \\ \hline
\multirow{4}{*}{subtask2\_track1} & \textbf{YNU-HPCC} & \textbf{96.6} & \textbf{93.6}  & \textbf{95.1}  \\
                                  & ZZU\_NLP          & 93.8 & 91.82 & 92.8  \\
                                  & ShaunTheSheep     & 92.2 & 92.81 & 92.5  \\
                                  & kangreen          & 92.8 & 91.8  & 92.3  \\ \hline 
subtask1\_track2                  & \textbf{YNU-HPCC} & \textbf{98.4} & \textbf{97.4}  & \textbf{97.9} \\ \hline
subtask2\_track2                  & \textbf{YNU-HPCC} & \textbf{95}   & \textbf{93.2}  & \textbf{94.1}  \\
\toprule[1.5pt]
\end{tabular}
\vspace{-2.0em}
\end{table}
Since this task contains two subtasks, each subtask can submit results according to the requirements of Track 1 and Track 2. Therefore, there are four results in total: subtask1\_track1, subtask1\_track2, subtask2\_track1, and subtask2\_track2. The experimental results of subtask 1\_track 1 are shown in Table~\ref{subtask1}, where the Human's language rule-based method identifies the commonalities and vehicles contained in metaphors based on human language habits. For example:
\begin{itemize}
    \item[$\bullet$] The tenor and vehicle in Chinese metaphorical sentences are usually connected by comparative words such as "\begin{CJK*}{UTF8}{gbsn}像\end{CJK*}" (like), "\begin{CJK*}{UTF8}{gbsn}如\end{CJK*}" (as), "\begin{CJK*}{UTF8}{gbsn}似\end{CJK*}" (seem), "\begin{CJK*}{UTF8}{gbsn}是\end{CJK*}" (be), etc.
    \item[$\bullet$] The tenor, vehicle, and ground often appear directly in the metaphor.
\end{itemize}

As shown in Table~\ref{subtask1}, both the Human language rules-based method and the fine-tuned DeBERTa model performed well on the validation set, particularly the DeBERTa-finetuned model. Thus, the DeBERTa-finetuned model's results will be used for the Track 2 submission. Note that the DeBERTa-related results in Table~\ref{subtask1} were obtained by splitting the validation set 80/20, training on 80\%, and predicting the remaining 20\%.
\begin{figure}[t!]
    \centering
    \begin{subfigure}[b]{0.45\textwidth}
        \centering
        \includegraphics[width=\textwidth]{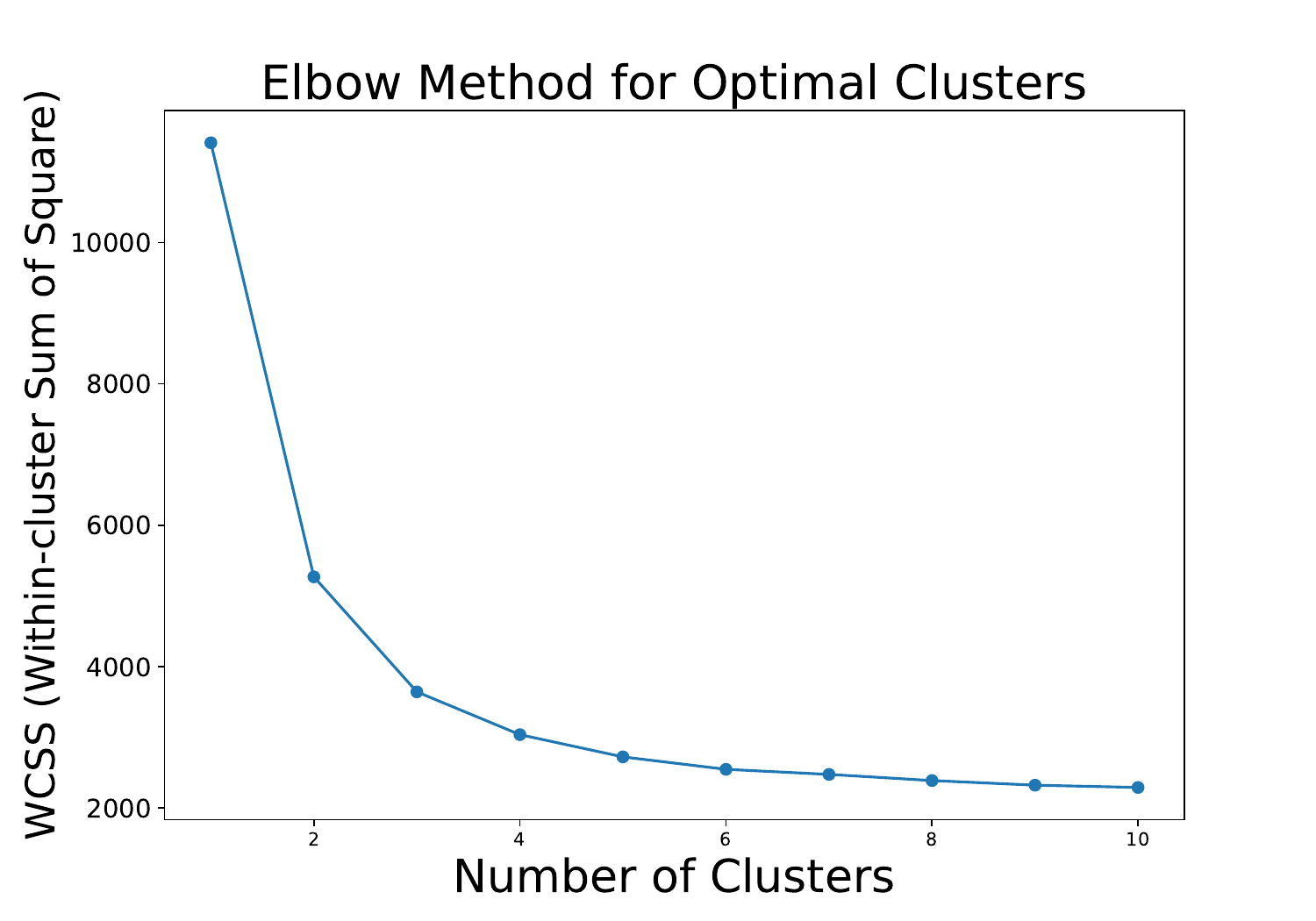}
        \caption{Select the optimal cluster using the elbow method for subtask 1.}
        \label{fig:sub1}
    \end{subfigure}
    \hfill
    \begin{subfigure}[b]{0.45\textwidth}
        \centering
        \includegraphics[width=\textwidth]{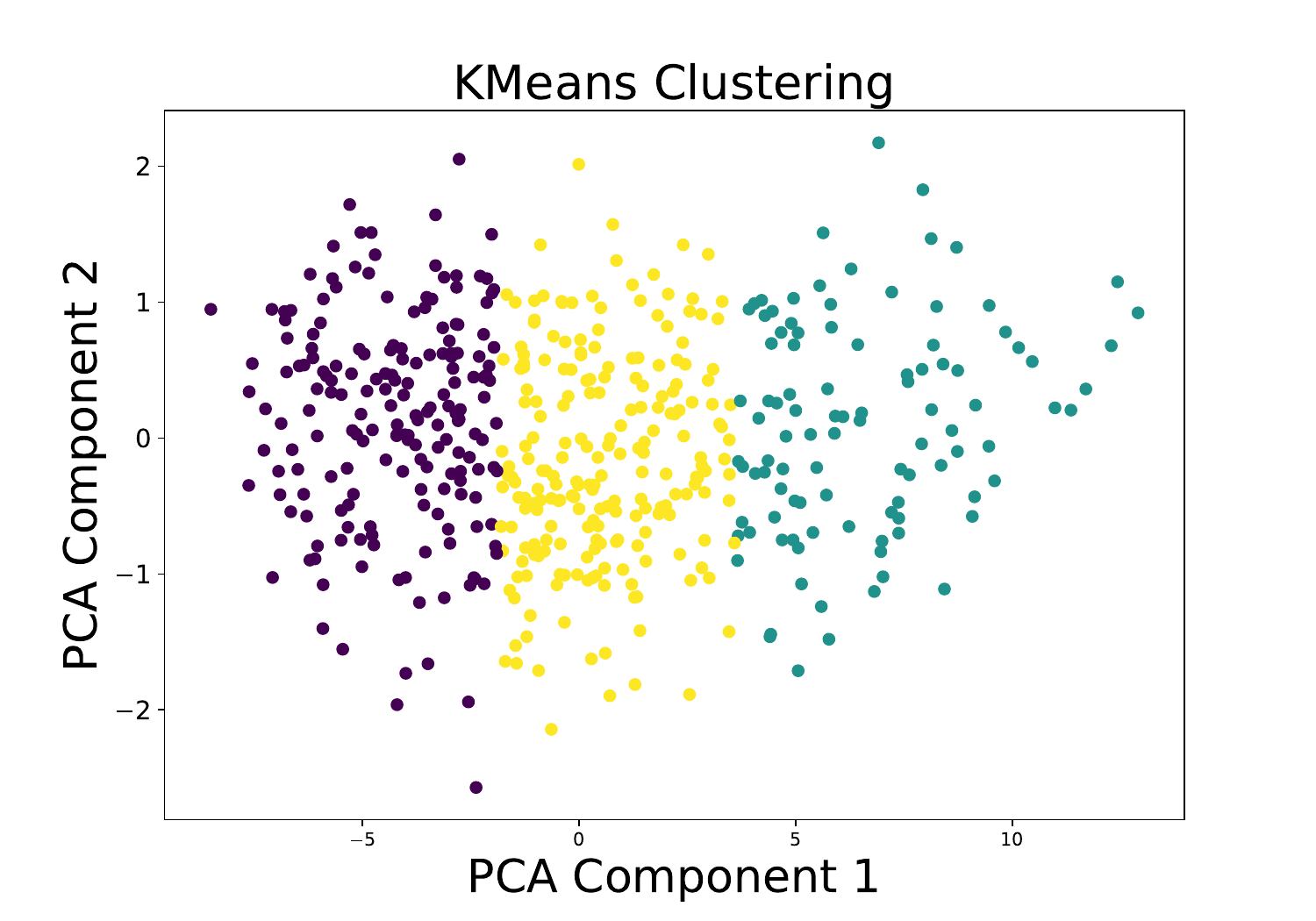}
        \caption{Scatter plot of clustering results based on PCA for subtask 1. }
        \label{fig:sub2}
    \end{subfigure}
    \vfill
    \begin{subfigure}[b]{0.45\textwidth}
        \centering
        \includegraphics[width=\textwidth]{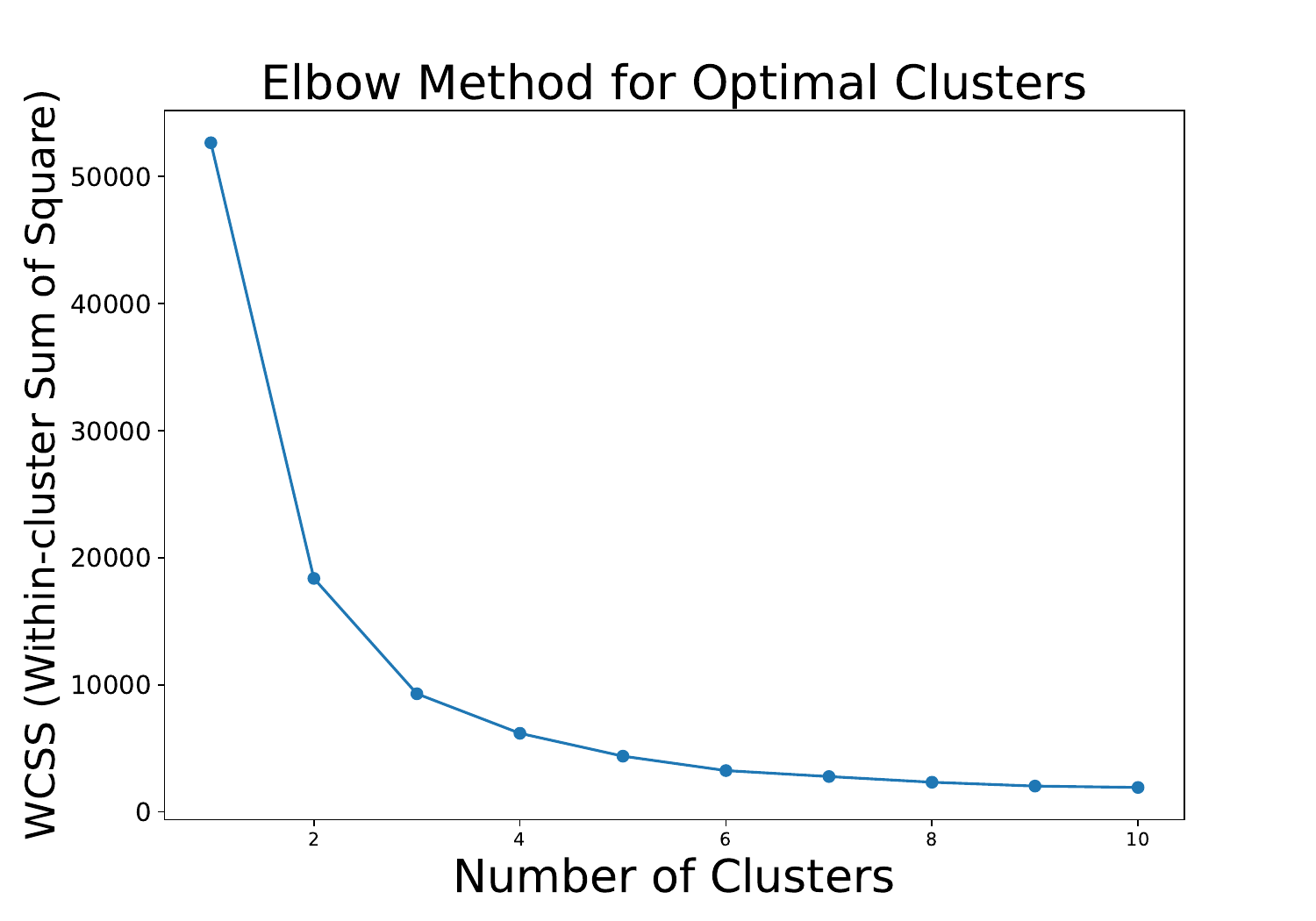}
        \caption{Select the optimal cluster using the elbow method for subtask 2.}
        \label{fig:sub3}
    \end{subfigure}
    \hfill
    \begin{subfigure}[b]{0.45\textwidth}
        \centering
        \includegraphics[width=\textwidth]{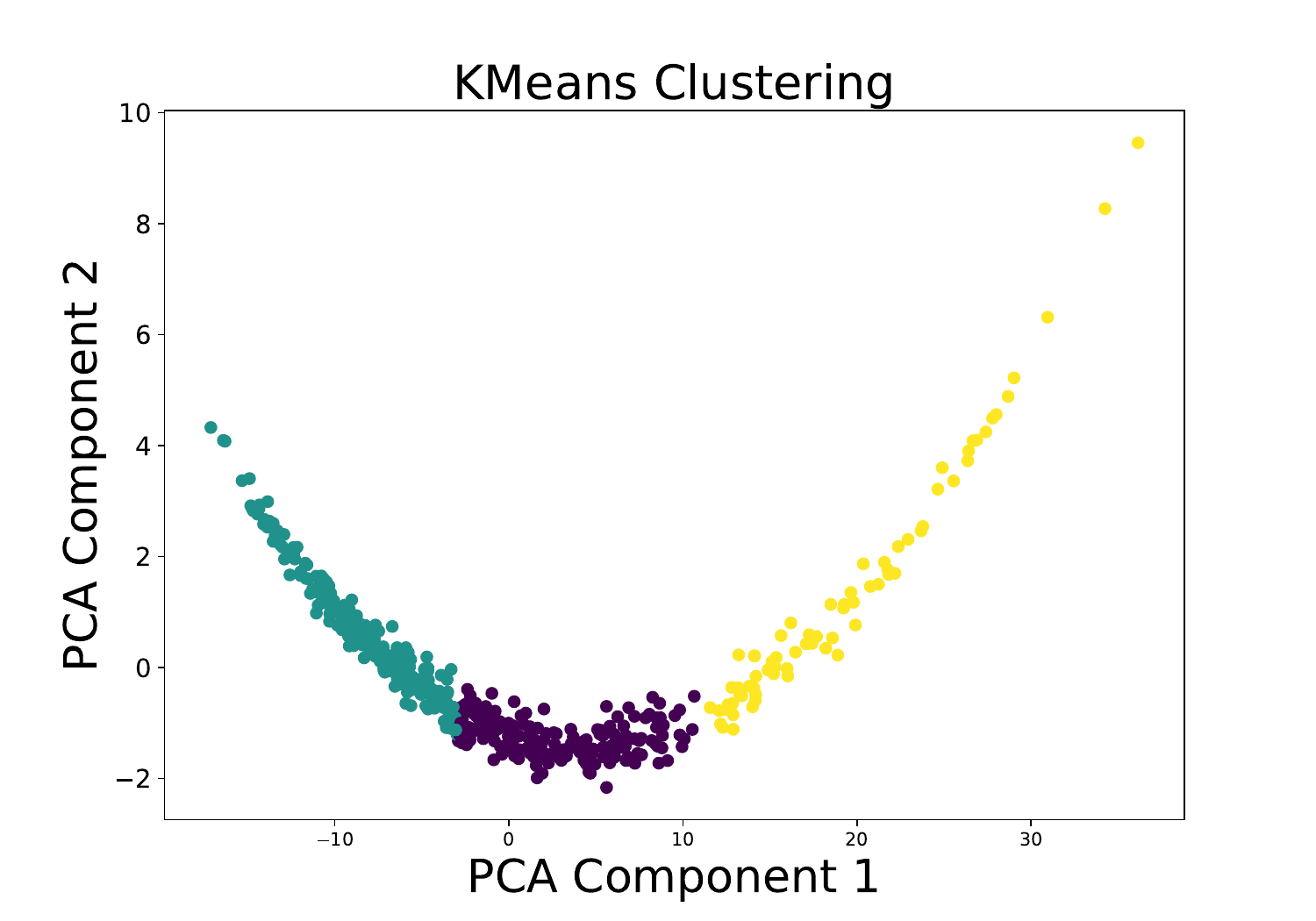}
        \caption{Scatter plot of clustering results based on PCA for subtask 2. }
        \label{fig:sub4}
    \end{subfigure}
    \caption{The results of k-means clustering experiment.}
    \label{fig: main}
\vspace{-2.0em}
\end{figure}

\begin{figure}[t!]
\centering
    \begin{subfigure}[b]{0.49\textwidth}
        \centering
        \includegraphics[width=\textwidth]{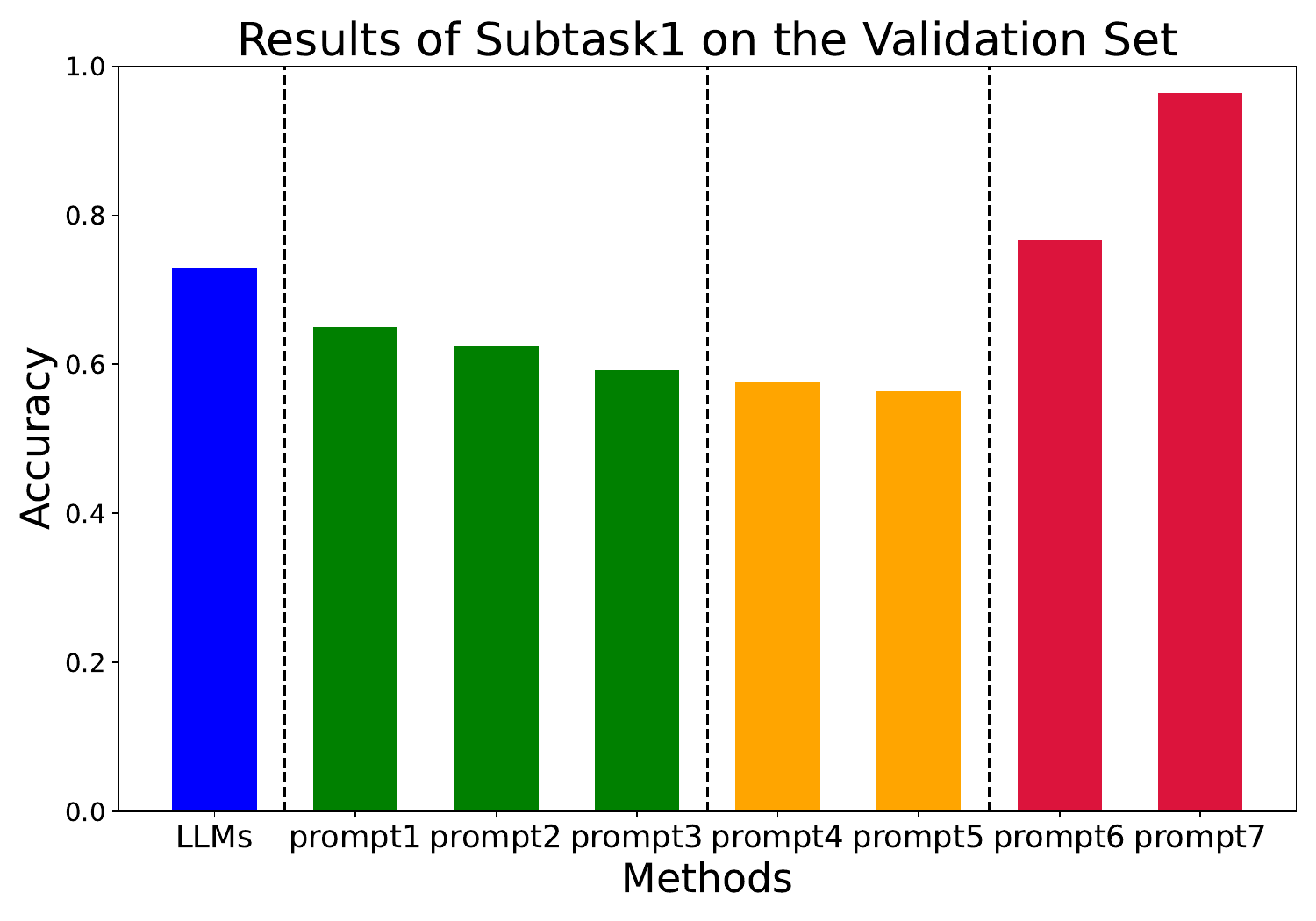}
        \caption{The impact of different prompts on LLMs in Subtask 1.}
        \label{accuracy-figure}
    \end{subfigure}
    \hfill
    \begin{subfigure}[b]{0.49\textwidth}
        \centering
        \includegraphics[width=\textwidth]{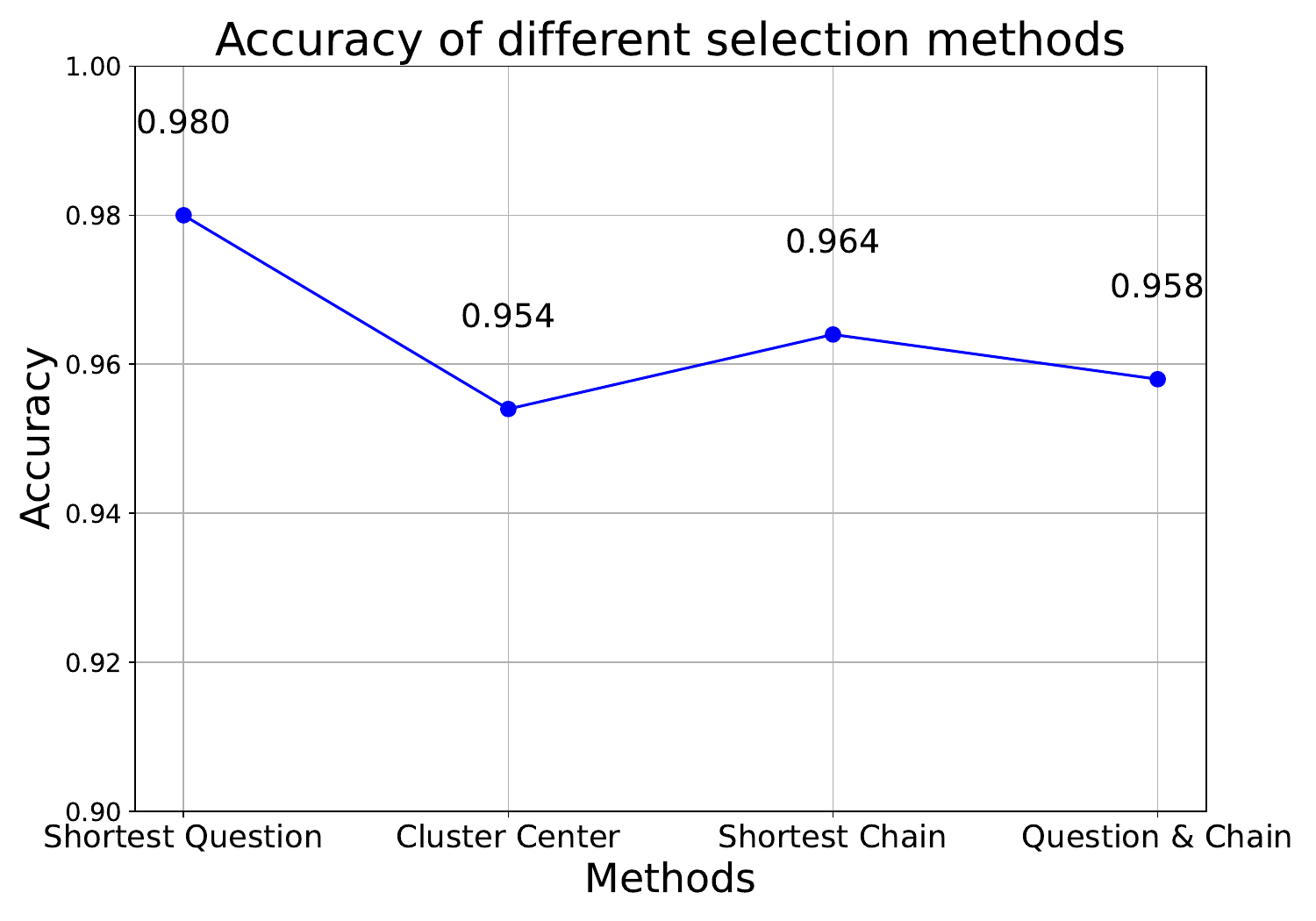}
        \caption{The impact of different selection methods on LLMs in Subtask 1.}
        \label{select}
    \end{subfigure}
    \caption{Accuracy of using different prompts (a) and different sampling rules for example sampling (b).}
\end{figure}
\begin{figure}[t!]
    \centering
    \includegraphics[width=7cm]{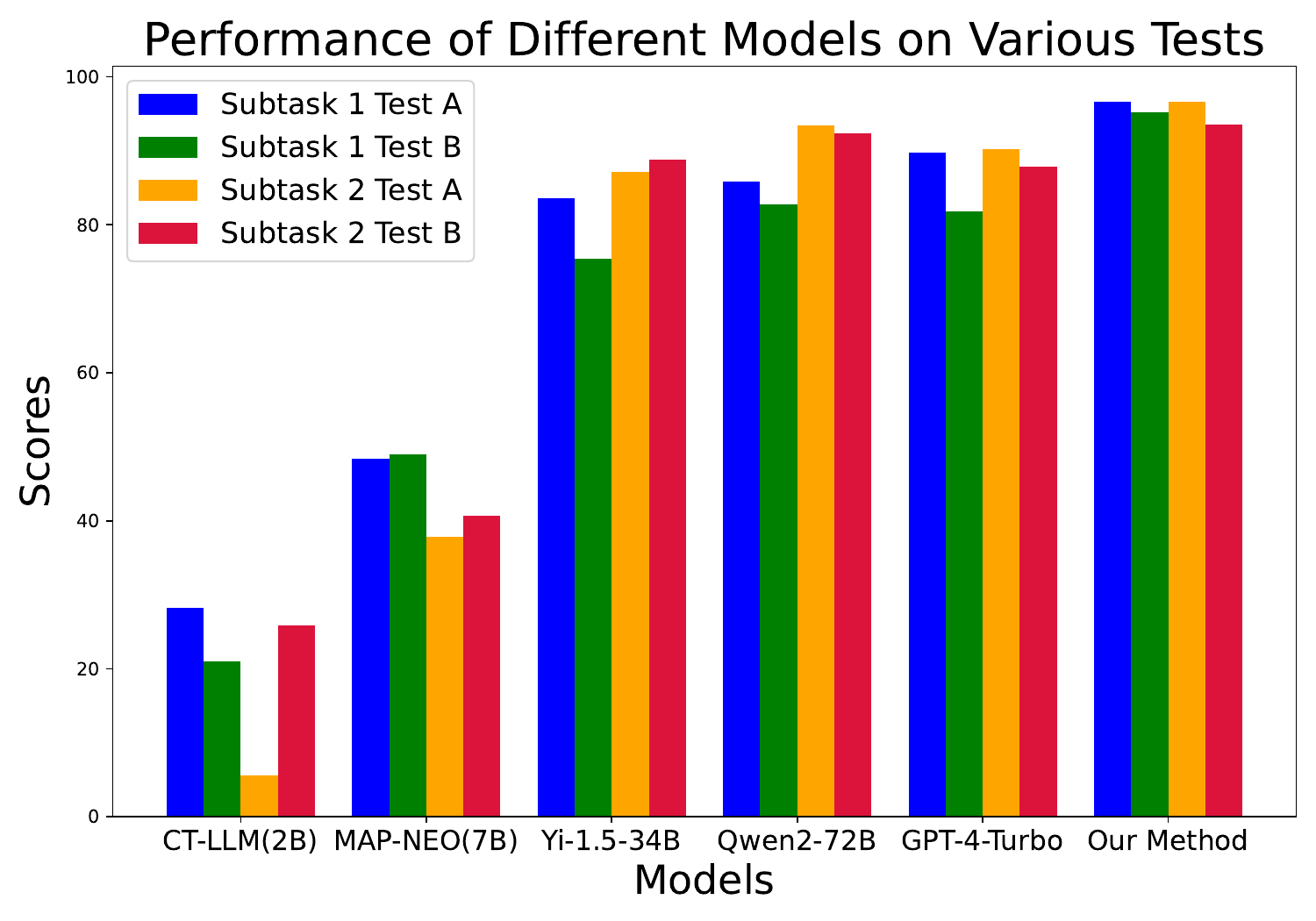}
    \caption{Compare the results with the benchmark model}
    \label{compare}
    \vspace{-2.0em}
\end{figure}
\subsection{Analysis and Discussion}
\subsubsection{Question Clustering and Demonstration Sampling.} 
When clustering the question using the k-means algorithm, the number of clusters was chosen as 3 based on the elbow method shown in Fig.~\ref{fig:sub1} and Fig.~\ref{fig:sub3}. The final clustering scatter plot is presented in Fig.~\ref{fig:sub2} and Fig.~\ref{fig:sub4}. Several methods of selecting examples have been tried, such as choosing the shortest length of the question, choosing the shortest length of the chain of thought, choosing the cluster center, and choosing the shortest length of both the question and the chain of thought. According to the experimental results shown in Fig.~\ref{select}, finally, the example with the shortest question length has been selected.
\vspace{-0.2em}
\subsubsection{Direct Response Provision to LLMs.} 
Initially, the answers were directly provided to the LLMs, such as Prompt 1, Prompt 2, and Prompt 5. Using Prompt 1 as an example: if DeBERTa predicts A for a question, this result is provided to LLMs as a reference, with a note that it may not be correct. Despite DeBERTa-finetuned answers and language rules showing good results, LLMs did not use the reference answers and their effectiveness significantly decreased as shown in Fig.~\ref{accuracy-figure}.
\vspace{-0.3em}
\subsubsection{Reference answers and explanations.}
Since directly providing reference answers to LLMs was ineffective, we added reasons for these answers to persuade the models. For example, in Prompt 3 and Prompt 4, the only difference between them and Prompt 1 by including LLMs results and their reasoning. However, this approach led to a further decrease in LLMs' effectiveness as shown in Fig.~\ref{accuracy-figure}.
\vspace{-0.3em}
\subsubsection{Answer Candidates.}
Using Prompt 7 as an example: LLMs can refer to the $\mathbf{p}$ generated by the DeBERTa-finetuned model. Experimental results show that answer candidates significantly enhance LLM effectiveness, even when DeBERTa's performance is poor. For instance, with DeBERTa-un\_finetune accuracy at 0.234, introducing $\mathbf{p}$ increased LLM effectiveness from 0.73 to 0.766.
\vspace{-1.0em}
\subsubsection{Ablation Analysis.}
As shown in Table~\ref{subtask1}, the effectiveness of LLMs has been dramatically improved by our proposed method, with an accuracy of 0.978 on the validation set. During ablation experiments, the generation of answer candidates had the most significant impact on our method. When there were no candidates, the accuracy was 0.906. When there are no demonstrations, the accuracy is 0.974, which is 0.1 higher than prompt 7's accuracy of 0.964 since "Let's think step by step." was added in the prompt.
\vspace{-0.4em}
\subsubsection{Official Results.}
The official provides two test sets, test A and test B. The official offers benchmark results for the MAP-NEO \cite{zhang2024map}, CT-LLM \cite{du2024chinese}, Yi-1.5-34B, Qwen2-72B, and GPT-4-Turbo models on the test set. From Fig.~\ref{compare}, it can be seen that our method has obvious advantages and more stable effects. Except for ranking third on subtask1\_track1, it ranks first on all other tasks as shown in Table~\ref{official_rusults}.
\begin{figure}[t!]
\centering 
\includegraphics[height=6cm]{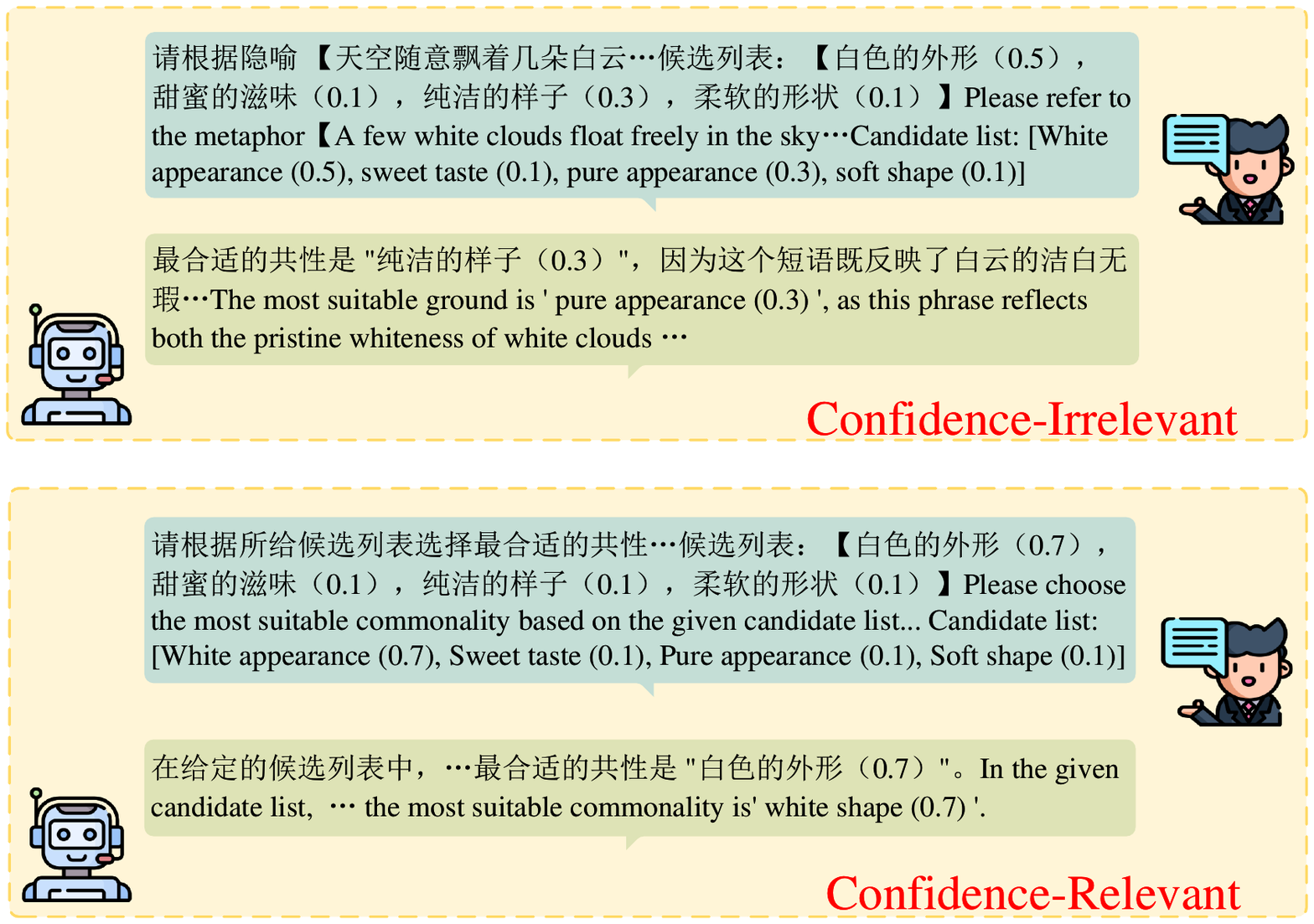}
\caption{Examples of Communicating with Large Language Models.} 
\label{example}
\vspace{-2.0em}
\end{figure}
\vspace{-0.4em}
\subsubsection{Discussion.}
Introducing a confidence score notably enhanced LLM performance, but we had further thoughts after its implementation. Fig.~\ref{example} shows the process of communicating with LLMs to identify correct ground, with examples from the training set, where the symbolic sentence is a poem: "\begin{CJK*}{UTF8}{gbsn}天空随意飘着几朵白云，棉花糖一样如你般的纯净，打开手机看到今天天气晴，如果有可能想带你去远行，某年某月某日几点零几。\end{CJK*}" (A few white clouds float freely in the sky, as pure as you, like cotton candy. When I open my phone and see that the weather is sunny today, I want to take you on a long journey if possible. What time is it in a certain year, month, or day), the tenor is "\begin{CJK*}{UTF8}{gbsn}白云\end{CJK*}" (white clouds), the vehicle is "\begin{CJK*}{UTF8}{gbsn}棉花糖\end{CJK*}" (cotton candy), and the correct ground is "\begin{CJK*}{UTF8}{gbsn}白色的外形\end{CJK*}" (white appearance). 

\noindent\textbf{Q: Do LLMs rely solely on confidence scores for decisions?} In other words, does the model only pick the highest confidence option without reasoning?

\noindent\textbf{A: The answer is negative.} According to the Confidence-Irrelevant section in Fig.~\ref{example}, it can be observed that the magnitude of the confidence score does not solely determine the LLMs' choice. For example, although the highest confidence score is for "\begin{CJK*}{UTF8}{gbsn}白色的外形\end{CJK*}" (white appearance), the model ultimately chose "\begin{CJK*}{UTF8}{gbsn}纯洁的样子\end{CJK*}" (pure appearance).

\noindent\textbf{Q: Under what circumstances does the model refer to the confidence scores to make a choice?}

\noindent\textbf{A: When there is a significant difference in confidence scores.} From the Confidence-Relevant section in Fig.~\ref{example}, when the confidence score for "\begin{CJK*}{UTF8}{gbsn}白色的外形\end{CJK*}" (white appearance) is much higher than for "\begin{CJK*}{UTF8}{gbsn}纯洁的样子\end{CJK*}" (pure appearance), the model will revise its choice to select "\begin{CJK*}{UTF8}{gbsn}白色的外形\end{CJK*}" (white appearance).

\section{Conclusion}
A multi-stage framework was proposed to effectively enhance the ability of LLMs to recognize grounds and vehicles in Chinese metaphorical sentences. In the first stage, the DeBERTa model is used to generate answer candidates. Experiments have shown that introducing answer candidates in the prompt improves the recognition performance of LLMs, and this effect persists even when DeBERTa performs poorly. In the second stage, problems in the validation set are clustered, and representative problems are selected as examples. By introducing answer candidates, a heuristic-enhanced prompt is formed. Experiments have demonstrated that this method effectively improves the capability of LLMs. Additionally, ablation experiments reveal that answer candidates and the generated examples significantly contribute to the final results.

\bibliographystyle{splncs04}
\bibliography{submission}
\end{document}